\documentclass[11pt,a4paper]{article}
\usepackage{authblk}
\usepackage[hyperref]{acl2020}
\usepackage{times}
\usepackage{latexsym}

\usepackage{gb4e}
\noautomath
\usepackage{epigraph}
\usepackage{tabularx}
\usepackage{multirow}
\usepackage{comment}
\usepackage{caption}
\usepackage{subcaption}
\usepackage{graphicx}
\usepackage{float}
\usepackage{placeins}
\usepackage{pifont}% http://ctan.org/pkg/pifont
\newcommand{\cmark}{\ding{51}}%
\usepackage{microtype}

\aclfinalcopy % Uncomment this line for the final submission
\def\aclpaperid{531} %  Enter the acl Paper ID here

\begin{document}

\title{Generating Counter Narratives against Online Hate Speech:\\ Data and Strategies}

\renewcommand\Authfont{\bfseries}

\author[1]{Serra Sinem Tekiro\u{g}lu}
\author[1,2]{Yi-Ling Chung}
\author[1]{Marco Guerini}

\affil[1]{Fondazione Bruno Kessler, Via Sommarive 18, Povo, Trento, Italy
\protect\\ \texttt{tekiroglu@fbk.eu,ychung@fbk.eu,guerini@fbk.eu}}
\affil[2]{University of Trento, Italy \protect\\ }

%\date{}

\maketitle
\begin{abstract}
Recently research has started focusing on avoiding undesired effects that come with content moderation, such as censorship and overblocking, when dealing with hatred online. The core idea is to directly intervene in the discussion with textual responses that are meant to counter the hate content and prevent it from further spreading. Accordingly, automation strategies, such as natural language generation, are beginning to be investigated. 
Still, they suffer from the lack of sufficient amount of quality data and tend to produce generic/repetitive responses. Being aware of the aforementioned limitations, we present a study on how to collect responses to hate effectively, employing large scale unsupervised language models such as GPT-2 for the generation of silver data, and the best annotation strategies/neural architectures that can be used for data filtering before expert validation/post-editing.

\end{abstract}

\section{Introduction}

% PARAGRAPH: Definition

%According to UNESCO Hate Speech refers to ``expressions that advocate incitement to harm based upon the targetÕs being identified with a certain social or demographic group" \cite{gagliardone2015countering}.
%Victims of hate speech are usually targeted because of various aspects such as gender, race, religion, sexual orientation, physical appearance. 

% My ultimate suggestion: Owing to the upsurge in the use of social media platforms over the past decade, Hate Speech became a pervasive issue by spreading quickly and widely.  
Owing to the upsurge in the use of social media platforms over the past decade, Hate Speech (HS) has become a pervasive issue by spreading quickly and widely.
%\textcolor{blue}{Owing to the pervasiveness of social media platforms in the past decade,
%\textcolor{blue}{Owing to nowadays pervasiveness of social media platforms,
%hate speech %this form of abusive language 
%can spread quickly and widely.
%At the same time
Meanwhile, it is difficult to track and control its diffusion, since nuances in cultures and languages make it difficult to %always 
provide a clear-cut distinction between hate and dangerous speeches \cite{schmidt2017survey}.
The standard approaches to prevent online hate spreading include the suspension of user accounts or deletion of hate comments from the social media platforms (SMPs), paving the way for the accusation of censorship and overblocking. Alternatively, to weigh the right to freedom of speech, shadow-banning has been put into use where the content/account is not deleted but hidden from SMP search results. Still we believe that we must overstep reactive identify-and-delete strategies, to responsively %proactively \textcolor{violet}{**Proactive means we are trying to prevent the problem before it happens, but we are not doing that..?} 
intervene in the conversations \cite{Bielefeldt2011,jurgens-etal-2019-just}.
%\textcolor{blue}{SMPs standard approaches to prevent hate spreading include} the suspension of user accounts or deletion of hate comments.
%\textcolor{blue}{Similar strategies that try to cope with the problem of censorship by} weighing the right to freedom of speech \textcolor{blue}{include the use of shadow-banning, where the content/account is not deleted but hidden from SMP search results. Still we believe that we must overstep reactive identify-and-delete strategies, to proactively intervene in the conversations \cite{Bielefeldt2011,jurgens-etal-2019-just}}. % also with the help of NLP tools %``\textit{the strategic response to hate speech is more speech}"
In this line of action, some Non-Govermental Organizations (NGOs) train operators to %to monitor SMPs and %\textcolor{blue}{intervene on SMPs, when hatred is detected, by writing so-called (***so called has a negative connotation!)}  
intervene in online hateful conversations by writing counter-narratives. A Counter-Narrative (CN) 
%(sometimes called counter-comment or counter-speech) 
is a non-aggressive response that offers feedback through fact-bound arguments and is considered as the most effective approach to withstand hate messages \cite{benesch2014countering, schieb2016governing}. %\textcolor{blue}{It is important to stress that operators are instructed and trained  to follow precise guidelines for writing CNs}, %. Such guidelines are highly consistent across languages and across NGOs, and 
%similar to those in `Get the Trolls Out' project\footnote{http://stoppinghate.getthetrollsout.org/}. These guidelines \textcolor{blue}{suggest} %tend to emphasize using fact-bound information and non-offensive language in order to avoid escalating the hatred in the discussion. 
To be effective, a CN should follow guidelines similar to those in `Get the Trolls Out' project\footnote{http://stoppinghate.getthetrollsout.org/}, %which emphasize using fact-bounded information and non-offensive language
in order to avoid escalating the hatred in the discussion.

Still, manual intervention against hate speech is not scalable. % given the incredible amount of hate continuously posted on SMPs: %Counter-narratives are the main instrument used by some NGOs} to tackle hatred online. 
%Automatizing the countering procedure \textcolor{blue}{can provide a viable solution.} %necessary to  increment  the  efficacy and effectiveness of hate countering.
Therefore, %data-driven NLG approaches are beginning to be investigated since they represent a promising solution in helping NGO operators 
data-driven NLG approaches are beginning to be investigated to assist NGO operators in writing CNs. %through automatic CN generation. 
As a necessary first step, diverse CN collection strategies have been proposed, each of which has its advantages and shortcomings \cite{mathew2018analyzing,qian-etal-2019-benchmark,conan-2019}.

In this study, we aim to investigate methods to obtain high quality CNs while reducing efforts from experts. We first compare data collection strategies depending on the two main requirements that datasets must meet: (i) data quantity and (ii) data quality. Finding the right trade-off between the two is in fact a key element for an effective automatic CN generation. To our understanding none of the collection strategies presented so far is able to fulfill this requirement. Thus, we test several \textit{hybrid} strategies to collect data, by mixing niche-sourcing, crowd-sourcing, and synthetic data generation obtained by fine-tuning deep neural architectures specifically developed for NLG tasks, such as GPT-2 \cite{radford2019language}. We propose using an author-reviewer framework in which an author is tasked with text generation and a reviewer can be a human or a classifier model that filters the produced output. %We evaluated our framework through 
Finally, a validation/post-editing phase is conducted with NGO operators over the filtered data. Our findings show that this framework is scalable allowing to obtain datasets that are suitable in terms of diversity, novelty, and quantity. %when limited resources are available

\section{Related Work} \label{related_work}

%\textcolor{red}{- NEW RELATED WORK GENERATION}

%Here 
We briefly focus on three research aspects related to hate online, i.e. available datasets, methodologies for detection, and studies on textual intervention effectiveness. In the next section we will instead focus on few methodologies specifically devoted to HS-CN pairs collection. 

\noindent\textbf{Hate datasets.} 
Several datasets have been collected from SMPs including Twitter \cite{waseem2016hateful, waseem2016you, ross2017measuring}, Facebook \cite{kumar2018benchmarking}, WhatsApp \cite{sprugnoli2018creating}, and forums \cite{de2018hate}, in order to perform hate speech classification \cite{xiang2012detecting, silva2016analyzing, del2017hate, mathew2018analyzing}.

\noindent\textbf{Hate detection.} Most of the research on hatred online focuses on hate speech detection \cite{warner2012detecting, silva2016analyzing, schmidt2017survey,fortuna2018survey} employing features such as lexical resources \cite{gitari2015lexicon, burnap2016us}, sentiment polarity \cite{burnap2015cyber} and multimodal information \cite{hosseinmardi2015detection} to a classifier. 

%Several works have investigated online English hate speech detection and the types of hate speech. Owing to the availability of current datasets, researchers often use supervised-approaches to tackle hate speech detection on SMPs including blogs \cite{warner2012detecting, djuric2015hate, gitari2015lexicon}, Twitter \cite{xiang2012detecting, silva2016analyzing, mathew2018analyzing}, Facebook \cite{del2017hate}, and Instagram \cite{zhong2016content}. The predominant approaches are to build a classifier trained on various features derived from lexical resources \cite{gitari2015lexicon, burnap2015cyber, burnap2016us}, n-grams \cite{sood2012automatic, nobata2016abusive} and knowledge base \cite{dinakar2012common}, or to utilize deep neural networks \cite{mehdad2016characters, badjatiya2017deep}. In addition, other approaches have been proposed to detect subcategories of hate speech such as anti-black \cite{kwok2013locate} and racist \cite{badjatiya2017deep}. \citet{silva2016analyzing} studied the prevalent hate categories and targets on Twitter and Whisper, but limited hate speech only to the form of \textit{I  $<$intensity$>$  $<$user intent$>$ $<$any word$>$}. A comprehensive overview of recent approaches on hate speech detection using NLP can be found in \cite{schmidt2017survey,fortuna2018survey}.
\noindent\textbf{Hate countering.} 
Recent work has proved that counter-narratives are effective in hate countering \cite{benesch2014countering, silverman2016impact, schieb2016governing, stroud2018varieties, mathew2019thou}. Several CN methods to counter hatred are outlined and tested by \citet{benesch2014countering},  \citet{munger2017tweetment}, and \citet{mathew2019thou}. %\citet{mathew2019thou} characterize the linguistic structure of comments on Youtube and observe that users favor comments covering counter-narratives in contrast to the ones covering hate speech.

\section{CN Collection Approaches} 
\label{aaa}
Three prototypical strategies to collect HS-CN pairs have been presented recently.
%In this section we focus on 
%There are three prototypical strategies %presented in three different papers 
%to collect HS-CN pairs, presented recently. %we will discuss their advantages. %To our knowledge these are the only works specifically addressing the problem of hate-speech counter narrative pairs collection.

\textbf{Crawling (CRAWL)}. \citet{mathew2018analyzing} focuses on the intuition that CNs can be found on SMPs as responses to hateful expressions.
%The work in \citet{mathew2018analyzing} focuses on the intuition that on SMPs, along with hate speeches we can also finds CNs as responses to such hate speeches. 
The proposed approach is a mix of automatic HS %content 
collection via linguistic patterns, and a manual annotation of replies to check if they are responses that counter the original hate content. Thus, all the material collected is made of \textit{natural/real} occurrences of HS-CN pairs.

\textbf{Crowdsourcing (CROWD)}. \citet{qian-etal-2019-benchmark} propose that
%The work present in \cite{qian-etal-2019-benchmark} focuses on the idea that,
once a list of HS is collected from SMPs and manually annotated, we can briefly instruct crowd-workers (non-expert) to write possible responses to such hate content. In this case the content is obtained in controlled settings as opposed to crawling approaches.

\textbf{Nichesourcing (NICHE)}. The study by  \citet{conan-2019} still relies on the idea of outsourcing and collecting CNs in controlled settings. However, in this case the CNs are written by NGO operators, i.e. persons specifically trained to fight online hatred via textual responses that can be considered as experts in CN production.
\section{Characteristics of the Datasets} 

Regardless of the HS-CN collection strategy, datasets %should always 
must meet two criteria: \textit{quality} and \textit{quantity}. While \textit{quantity} has a straightforward interpretation, %in the CN scenario 
we %believe
propose that data \textit{quality} should be decomposed into \textit{conformity} (to NGOs guidelines) and \textit{diversity} (lexical \& semantic). %(i.e. providing several and diverse kind of CN types and arguments).
Additionally, HS-CN datasets should not be ephemeral, which %or prone to data loss: this
is a structural problem with crawled data since, due to  copyright limitations, datasets are usually distributed as a list of tweet IDs %- especially in HS scenarios 
\cite{Klubicka2018}. With generated data (crowdsourcing or nichesourcing) the problem is avoided. 

\noindent\textbf{Quantity.} While the CRAWL dataset is very small and ephemeral, representing more a proof of concept than an actual dataset, the CROWD dataset involved more than $900$ workers to produce $\approx41K$ CNs, while the NICHE dataset is constructed by the participation of $\approx100$ expert-operators to obtain $\approx4K$ pairs (in three languages) and resorted to HS paraphrasing and pair translation to obtain the final $\approx14K$ HS-CN pairs. % As can be seen 
Evidently, employing non-experts, e.g, crowdworkers or %other type of
annotators, is preferable in terms of data quantity.

\noindent\textbf{Quality.} In terms of quality, %we first need to focus on diversity. In our view
we consider that diversity is of paramount importance, since verbatim repetition of arguments can become detrimental for operator credibility and for the CN intervention itself. % since we assume that the CN intervention is happening on a SMP - in a one to many conversation, not one to one - so verbatim repetition of arguments can soon become detrimental for operator credibility and for the discussion itself. 
Following %the intuition presented in
\citet{li-etal-2016-diversity}, we distinguish between (i) \textit{lexical diversity} and (ii)
\textit{semantic diversity}. While lexical diversity %basically 
focuses on the diversity in surface realization of CNs and can be captured by word overlapping metrics, %there is still the problem of capturing semantic diversity, in fact there can be CNs that are similar in meaning even if expressed with different wordings \textcolor{red}{(e.g.,``\textit{I don't know}" vs. ``\textit{I haven't a clue}")}.
semantic diversity focuses on meaning and is harder to be captured, as in the case of CNs with similar meaning but different wordings (e.g.,``\textit{Any source?}" vs. ``\textit{Do you have a link?}").

\noindent\textbf{(i) Semantic Diversity \& Conformity}. To model %these concepts
semantic diversity and conformity, we focus on the CN `argument' types that are present in various datasets. %This is a good method to understand
Argument types are useful in assessing content richness \cite{hua2019argument}. %- to understand whether CNs are generic or specific """NOTE: we did not say anything about genericness or specificness before,""" if they provides diverse information and knowledge when replying to HS.
In a preliminary analysis, CROWD CNs are observed to be simpler and mainly focus on `denouncing' the use of profanity while NICHE CNs are found richer with a higher variety of arguments. On the other hand, CRAWL CNs can cover diverse arguments to a certain extent while being highly prone to contain profanities. %instead, %while quite diverse in term of arguments (even if simpler than NICHE), is more prone to the use of profanities in CNs.
%In order to have 
To perform a quantitative comparison, we randomly %sub-sampled
sampled 100 pairs from each dataset and annotated them according to the CN types presented by \citet{benesch2016}, which is the most comprehensive CN schema. %which is the most used and covers more CN type cases.  
%Since CRAWL, NICHE and CROWD do not use the same annotation schema to define CN types, we decide to resort to the one used in \cite{benesch2016} because it is the most used and covers more CN type cases. In order to grant the same procedure for the three datasets, we  randomly sub-sampled  100 pairs from each dataset and annotated them according to the aforementioned schema.  

\begin{table}[h!]
\centering
\begin{tabular}{lccc}%{lrrr}p{7.5cm}
     \hline
     & CRAWL   & CROWD & NICHE\\
     \hline
Hostile & 50 & 0 & 0\\     
Denouncing & 16 & 76 & 10\\
Den.$+$Oth. & 0 & 10 & 9 \\
Other & 34 & 14 & 81\\
\hline
RR & 3.16 & 4.83 & 2.72\\ % WW_islam 6.77
\hline
\end{tabular}
\caption{Diversity analysis of the three datasets. Semantic diversity is reported in terms of CN type percentages, %s coverage (percentage), 
Lexical diversity in terms of Repetition Rate (RR - average over 5 shuffles).}
  \label{table:WWvsCONAN}
\end{table}

The results are reported in Table \ref{table:WWvsCONAN}. For the sake of conciseness we focus on the \textit{hostile}, \textit{denouncing} and \textit{consequences} classes, giving \textit{other} to all remaining types (including the \textit{fact} class). Clearly, CRAWL does not meet the conformity standards of CNs considering the vast amount of \textit{hostile} responses (50\%), still granting a certain amount of type variety (\textit{other:} 34\%).
%As can be seen, CRAWL, while granting a certain amount of variety in responses (\textit{other} class 34\%), it is dramatically affected (50\%) by a vast amount of very \textit{hostile} kind of responses that clearly do not meet conformity standards of counter narratives. 
Contrarily, CROWD conforms to the CN standards (\textit{hostile:} 0\%), %but it is more 
yet mostly focuses on pure \textit{denouncing} (76$\%$) or denouncing with simple arguments (10$\%$). The class \textit{other} (14$\%$) consists of almost only simple arguments, such as ``\textit{All religions deserve tolerance}". In NICHE instead, arguments are generally and expectedly more complex and articulated, %(this is something that can be expected since arguments were written by expert operators)
and represent the vast majority of cases (81\%). %Only a minority of the cases was concerned with denouncing alone ($10\%$). 
Few examples of CN types are given in Table \ref{table:examples}. %NB: As a final note on conformity only a small 2$\%$ of cases were found to be non conforming in WW.

\begin{table*}[h!]
\centering
\begin{tabular}{l|p{13cm}}%{lrrr}
     \hline
Hostile & ``Hell is where u belong! Stupid f***t... go hang yourself!!" \\
     \hline
Denouncing & ``The N word is unacceptable. Please refrain from future use."\\
     \hline
%Fact & ``Muslims have contributed much to the UK due to their sterling educational achievements. More than half of Muslim schools in England surpass the national percentage GCSE average of 5 or more GCSE's or equivalent A*-C grades including English and Math GCSEs. This enables Muslims to work in important fields of Medicine and Law amongst others to enrich the lives of UK citizens"\\
Fact & ``The majority of sexual assaults are committed by a family member, friend, or partner of the victim, and only 12\% of convicted rapists are Muslim. It is not the religion, its the individuals, whether they're Muslim or not."\\
\hline

%    \hline
\end{tabular}
\caption{Some examples of the categories relevant to our analysis. Hostile from CRAWL dataset, Denouncting from CROWD, Fact (other) from NICHE.}
  \label{table:examples}
\end{table*}

\noindent\textbf{(ii) Lexical Diversity}. The Repetition Rate (RR) is used to measure the repetitiveness of a collection of texts, by considering the rate of non-singleton n-gram types it contains %-- usually from unigrams to fourgrams --
\cite{cettolo2014repetition,bertoldi2013cache}. 
% The advantage of this metric over the count of distinct ngrams \cite{xu-etal-2018-diversity,li-etal-2016-diversity,richards1987type} %or over the Type/Token Ratios metric \cite{richards1987type} 
%is that it is not affected by set sizes %, since RR is computed 
%by averaging the statistics collected on a sliding window of 1000 words. This allows us to compare corpora of diverse sizes.
We utilize RR instead of the simple count of distinct ngrams \cite{xu-etal-2018-diversity,li-etal-2016-diversity} or the standard type/token ratio \cite{richards1987type} since it allows us to compare corpora of diverse sizes by averaging the statistics collected on a sliding window of 1000 words.
Since CROWD and NICHE contain repeated CNs for different HSs\footnote{while this is an explicit data augmentation choice in NICHE, for CROWD it seems to derive from writing the same CNs for similar HSs by crowd-workers.}, we first removed repeated CNs and then applied a shuffling procedure to avoid that CNs that are answering to the same HS (so more likely to contain repetitions) appear close together. Results in Table \ref{table:WWvsCONAN} show that NICHE is the dataset with more lexical diversity (lower RR), followed by CRAWL and CROWD.

\noindent\textbf{Discussion.} %From this preliminary analysis
We can reasonably conclude that: (i) crawling,  %, even if promising from some perspectives, it
as presented in \cite{mathew2018analyzing}, is not a mature procedure yet for CN collection, even if it is promising, % - at least as presented in \cite{mathew2018analyzing}. 
(ii) nichesourcing is the one producing the best and most diverse material by far, however it is also the most challenging to implement considering the difficulty of making agreements with NGOs specialized in CN creation %(it can be difficult to find and make agreements with NGO specialized in CN creation for doing data collection) 
and it does not provide sufficient amount of data. (iii) On the contrary, CROWD seems the only one that can grant the amount of data that is needed for deep learning approaches, but contains more simple and stereotyped arguments. A summary of the pros and cons of each collection approach is presented in Table \ref{table:data_coll_pros_cons}.

\begin{table}[h!]
\centering
\begin{tabular}{l|c|cc|c}%{lrrr}p{7.5cm}
     \hline
     & Quantity &\multicolumn{2}{c|}{Quality}& non-eph.\\
     &  & Conf. & Diver. &  \\
     \hline
     Crawl & \cmark & - & \cmark &  - \\
     Crowd. & \cmark & \cmark & - & \cmark \\
     Niche. & - & \cmark & \cmark & \cmark \\
     \hline
\end{tabular}
\caption{Comparison of different approaches proposed in the literature according to the main characteristics required for the dataset.}
  \label{table:data_coll_pros_cons}
\end{table}

%Since none of the aforementioned approaches alone can be decisive for creating proper CN datasets, in the following sections we %first inspect generation capabilities of some SOTA generation models and then 
%propose a novel framework %where we try to understand if we can
%that mixes the output of sub-optimal generation models with non-expert and expert annotation to reduce data collection effort. %by leveraging the pros of each approach.

%\section{Data Augmentation using Author-Reviewer strategies}
\section{CN Generation through Author-Reviewer Architecture}
%******NOTE***** We have to find a meaningful section title that is also consistent with the previous section titles: "Data Collection Mixed Strategies: Human in the loop with Author-Reviewer Generation Approaches"
Since none of the aforementioned approaches alone can be decisive for creating proper CN datasets, we propose a novel framework that %mixes the output of sub-optimal generation models with non-expert and expert annotation to reduce data collection effort. Our main goal is to design and test approaches that 
mixes crowdsourcing and nichesourcing %in different ways for 
to obtain new quality data while reducing collection cost/effort. The key elements of this mix are: 
(i) There must be an external element in this framework that produces HS-CN candidates, (ii) Non-experts should pre-filter the material to be presented/validated by experts. %In fact, non-expert might not be able to produce high-quality CNs but they are still able to recognize them when reading. 
 %Of course neither the expert nor the non-expert can play this role, otherwise we would fall back to the standard crowd/niche-sourcing configurations.
Thus, we settle on the author-reviewer modular architecture \cite{oberlander2000stochastic,manurung-etal-2008-implementation}. 
In this architecture the author has the task of generating a text that conveys the correct
propositional content (a CN), whereas the reviewer must ensure that the authorÕs output satisfies certain quality properties. %whatever macroscopic properties have been imposed on it. 
The reviewer finally %ranks on : means attacking someone
evaluates the text viability and picks the ones to present to the NGO operators for final validation/post-editing.

%\textcolor{red}{
The author-reviewer architecture that we propose differs from the previous studies in two respects: (i) it is used for data collection rather than for NLG, (ii) we modified the original configuration by adding a human ÒreviewerÓ and a final post-editing step. 
%Also the evaluation metrics used are different, e.g. rather than focusing on fluency - as in Stochastic text generation by Oberlander & Brew - we focused on RR and Novelty metrics.
%"DO WE HAVE TO ADD THIS? SEEMS A BIT FAR STRETCHED TO FIND A DIFFERENCE": Moreover, rather than focusing on the fluency of the generated text \cite{oberlander2000stochastic}, we evaluated the generated text using Repetition Rate and Novelty metrics. 
%}

\begin{figure}
 \centering
\includegraphics[width=.79\linewidth]{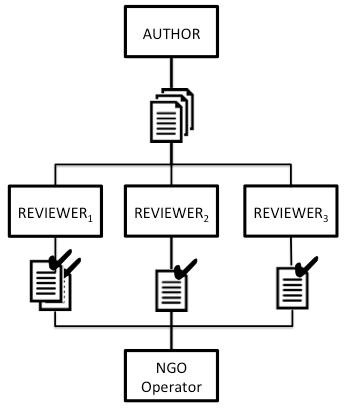}
\caption{The author-reviewer configuration. The author module produces HS-CN candidates and the reviewer(s) filter them. Finally, an NGO operator validates and eventually post-edits the filtered candidates.}
 \label{fig:author-reviewer}
\end{figure}

We first tested four different author configurations, then three reviewer configurations keeping the best author configuration constant. A representation of the architecture is shown in Figure \ref{fig:author-reviewer}.

%\noindent\textbf{(A1)} \textbf{Transf.+NICHE}: the author is a a transformer model trained on NICHE. \\
%\textbf{(A2)} \textbf{Transf.+CROWD}: the author is a a transformer model trained on CROWD. \\
%\textbf{(A3)} \textbf{GPT-2+NICHE}: the author is a a pre-trained model fine-tuned on NICHE. \\
%\textbf{(A4)} \textbf{GPT-2+NICHE}: the author is a a pre-trained model fine-tuned on CROWD.\\
%\textcolor{red}{- describe the methodology with the students (give scores from 0 to 3) how much instruction (half an hour with 20/30 examples of training).}\\

%\textcolor{red}{SERRA: the filtering module, rather than being optimized on standard metrics such as accuracy or F1 is optimized on false positive reduction. In fact, in our scenario since we have an overgeneration stage (GPT-2) and we want to save Operators effort, the best strategy is to keep false positive at a minimum, also at the cost of some false negative rate increase. Said differently: since we do not have a problem in producing examples, we prefer to trow away some good examples rather than retaining negative ones.}\\
\section{The Author: Generation Approaches}

%%%%%% NOTE:  we can also briefly describe suggestion based on IR model if we want
%Although carefully curated hate speech/counter-narrative datasets through, the data quantity still can be a drawback.
%\textcolor{violet}{It is not so important to mention that they did those models here. Lets put what we did.}
%In \cite{qian-etal-2019-benchmark} few baseline neural models for CN generation models were proposed, but the fact that these models were not pre-trained represent a limit to output grammatically and/or diversity given either the limited amount of training data (NICHE) or its repetitiveness (CROWD).

In order to obtain competent models that can provide automatic counter-narrative hints and suggestions to NGO operators, we have to overcome data bottleneck/limitations, i.e. either the limited amount of training data in NICHE or its repetitiveness in CROWD, especially for using neural NLP approaches. %PARAPHRASE: LM pre-training is helpful when fine-tuned for difficult generation tasks like chit-chat dialog and dialogbased question answering systems as well (Wolf et al., 2019) (Dinan et al., 2018). 
Pre-trained Language Models (LMs) have achieved promising results when fine-tuned on challenging generation tasks such as chit-chat dialog \cite{wolf2019transfertransfo, golovanov2019large}. To this respect, we propose using the recent large-scale unsupervised language model %GPT and 
GPT-2 \cite{radford2019language}, %, released by OPENAI, 
which is capable of generating coherent text and can be fine-tuned and/or conditioned on various NLG tasks. It is a large transformer-based \cite{vaswani2017attention} LM trained on a dataset of 8 million web pages. We used the medium model, which was the largest available during our experimentation and contains 345 million parameters, with 24 layers, 16 attention heads, and embedding size of 1024. %The model is a Transformer \cite{vaswani2017attention} with 24 layers, 16 attention heads, and embedding size of 1024. %Therefore, while keeping the advantages of either nichesourcing or crowdsourcing, \textcolor{red}{[SERRA?] we could see if these models are powerful enough to generate \textit{coherent and diverse} CN suggestions to human operators.\\ % which provides the benefits of quality and resistance to data loss,
%In the following experiment we will still use non-pretrained model for comparison purposes to undestand the advantages/benefits of pre-trained models.} 
We fine-tuned two models with GPT-2, one on NICHE and one on CROWD datasets, %obtaining 2 models 
for counter-narrative generation.% given a HS as input. 
%https://d4mucfpksywv.cloudfront.net/papers/GPT_2_Report.pdf

\textbf{NICHE - Training and test data.} %The data split we used includes 
We have split 5366 pairs of HS-CN for training and the rest (1288 pairs) for testing. %We make sure that no hate speech was present in training at test simultaneously.
In particular, the original HS-CN pairs, one HS paraphrase, and the pairs translated from FR and IT were kept for training while the other HS paraphrases were used for testing. See \citet{conan-2019} for further details.
%In particular, the split has been done by keeping one HS paraphrase for testing while the original HS-CN pairs and the other HS paraphrase were included in training. We also included the pairs translated from FR and IT - see \citet{conan-2019} for further details. 

\textbf{CROWD - Training and test data.} Although the CROWD dataset was created for dialogue level HS-CN, we could extract HS-CN pairs by selecting the dialogues in which only 1 utterance was labeled as HS. Therefore, we could guarantee that the crowd-produced CNs are exactly for the labeled utterance. We then applied a 80/20 training and test split, obtaining 26320 and 6337 pairs. % Similarly to the NICHE configuration we split the data with respect to the unique HSs. %, we make sure that no hate speech was present in training at test simultaneously, so

\noindent\textbf{Generation Models.} We fine-tuned GPT-2\footnote{We adopted the fine-tuning implementation from \url{https://github.com/nshepperd/gpt-2}}, with a batch size of 1024 tokens and a learning rate of 2e-5. The training pairs %for each dataset 
are represented as $[HS\_start\_token]\;HS\; [HS\_end\_token]$ $[CN\_start\_token]\;CN\; [CN\_end\_token]$. While we empirically selected model checkpoint at the $3600^{th}$ step of fine-tuning with NICHE dataset, with CROWD dataset we selected the checkpoint at the $5000^{th}$ step. After fine-tuning the models % with HS-CN data, 
the generation of CNs for the test HSs has been performed using Nucleus Sampling \cite{holtzman2019curious} with a p value of 0.9, which provides an enhanced diversity on the generation in comparison to the likelihood maximization decoding methods while preserving the coherency by truncating the less reliable tail of the distribution. At the test time, the input HSs are fed into models as conditions, which are used as the initial contexts while sampling the next tokens. Given an input HS, %GPT-2 fine-tuned 
the models produce a chunk of text which is a list of HS-CN pairs of which the first sequence marked with $[CN\_start\_token]\;CN\; [CN\_end\_token]$ is the generated output.% On the other hand, we observed that the rest of the generated output consists of semantically coherent brand-new HS-CN pairs, marked with proper HS/CN start and end tokens consistent with the training data representation. 

\noindent\textbf{Baselines.} %On the other hand, we need to evaluate the generation with our data only instead of using these huge models. 
In addition to the fine-tuned GPT-2 models, % models that we obtained by , %the pretrained transformer-based LM's \textcolor{purple}{[Would it be better to directly use GPT-2 instead of the pretrained transformer-based LM's?]}, 
we also evaluate two baseline models. Considering the benefits of transformer architectures on parallelization and learning long-term dependencies over recurrent models \cite{vaswani2017attention}, we have implemented the baseline models using transformer architecture. The models have been trained similar to the base model described by \citet{vaswani2017attention} with 6 transformer layers, batch size of 64, 100 epochs, 4000 warmup steps, input/output dimension of 512, 8 attention heads, inner-layer dimension of 2048, drop-out rate of 0.1. %For NICHE  model, we trained it 8381 steps and for CROWD ** steps corresponding to 100 epochs for each. 
%Similar to the fine-tuned models,
We used Nucleus Sampling also for the baselines with a p value of 0.9 during decoding. 
% WE HAVE TO WRITE adam optimizer if accepted. 
\begin{table}[t!]
\centering
\begin{tabular}{lcccc}%{lrrr}p{7.5cm}
     \hline
     Author & RR & Novel. & BLEU & BertS. \\
     \hline
    TRF$_{crowd}$ & 8.93 & 0.04 & 0.305 & 0.485\\ 
    GPT$_{crowd}$  & 5.89 & 0.46 & 0.270 & 0.482\\
     \hline
    TRF$_{niche}$  & 4.89 & 0.10 & 0.569 & 0.457 \\
    GPT$_{niche}$    & 3.23 & 0.70 & 0.316 & 0.445\\
     \hline
\end{tabular}
\caption{Evaluation results of best author configuration with different datasets. Novelty is computed w.r.t. to the corresponding training set, RR in the produced test output.}
  \label{table:generation_results}
\end{table}

%\textcolor{red}{
In brief, we have trained four different configurations/models as authors:
\begin{enumerate}
\itemsep0em 
\item \textbf{TRF$_{crowd}$}: baseline on CROWD dataset
\item \textbf{GPT$_{crowd}$}: fine-tuned GPT-2 on CROWD dataset
\item \textbf{TRF$_{niche}$}: baseline on NICHE dataset 
\item \textbf{GPT$_{niche}$}: fine-tuned GPT-2 on NICHE dataset
\end{enumerate}
%}

\noindent\textbf{Metrics.} %To evaluate the various models, 
We report both standard metrics (BLEU \cite{papineni2002bleu}, BertScore \cite{zhang2019bertscore}) concerning the lexical and semantic generation performances 
and a specific \textit{Diversity} metric (RR) regarding the generation quality. As a second quality metric, we report \textit{Novelty} \cite{wang2018sentigan} based on Jaccard similarity function (a variant of the same metric is used also by \citet{dziri2018augmenting}). 
While diversity is used to measure the ability of the model to produce diverse/varied responses with respect to the given input HS, % that are varied according to the varying of the input HS, 
novelty is used to measure how different the generated sequences are with regard to the training corpus \cite{wang2018sentigan}.

%In other words, we want to see if the generator simply copies the sentence in the corpus instead of generating new ones.

\noindent\textbf{Results.} Results of the author model experiments are shown in Table \ref{table:generation_results}. In terms of BLEU and BertScore, baseline models yield a better performance. %However, it is crucial to mention that we cannot directly and objectively compare the presented scores among the models.
However, a few peculiarities of CN generation task and the experiment settings hinder the direct and objective comparison of the presented scores among the models. First, gathering a finite set of all possible counter-narratives for a given hate speech is a highly unrealistic target. Therefore, we have only a sample of %possible and
proper CNs for each HS, which is a possible explanation of very low scores using the standard metrics. Second, the train-test splits of NICHE dataset contain same CNs since the splitting has been done using one paraphrase for each HS and its all original CNs, while CROWD train-test splits have a similar property since an exact same CN can be found for many different HSs. Consequently, the non-pretrained transformer models, which are more prone to generating an exact sequence of text from the training set, show a relatively better performance with the standard metrics in comparison to the advanced pre-trained models. Some randomly sampled CNs, generated by the various author configurations, are provided in Appendix.

%Considering generation quality, when we compare pre-trained vs. transformer approaches we see  GPT-2 ability to add a notable variability (RR reduction) that the non pre-trained models cannot achieve given either the small training sets (NICHE), 4.89 vs 3.23, or the small training data variability, 8.93 vs. 5.89, (CROWD). Moreover, GPT-2 provides an impressive boost in novelty (0.04 vs. 0.40 and 0.10 vs. 0.70).
%If we turn to generation quality, 
Regarding the generation quality, we observe that baseline models cannot achieve the diversity achieved by GPT-2 models in terms of RR -- both for NICHE and CROWD (4.89 vs 3.23, and 8.93 vs. 5.89). Moreover, GPT-2 provides an impressive boost in novelty (0.04 vs 0.46 and 0.10 vs 0.70).
Among the GPT-2 models, %we observe that 
the quality scores (in terms of RR and novelty) of the CNs generated by GPT$_{niche}$ are more than double in comparison to those generated with GPT$_{crowd}$.

With regard to the overall results, GPT$_{niche}$ 
is the most promising configuration to be 
employed as author. % Incidentally, 
In fact, we observed that, after the output CN, the over-generated chunk of text consists of semantically coherent brand-new HS-CN pairs, marked with proper HS/CN start and end tokens consistent with the training data representation. Therefore, on top of CN generation for a given HS, we can also take advantage of the over-generation capabilities of GPT-2, so that the author module can continuously output plausible HS-CN pairs without the need to provide the HS to generate the CN response. This expedient allows us to avoid the ephemerality problem for HS collection as well. 
%NOTE TO MARCO: We should talk about ephemerality, letting people know that data gets loss because of ID sharing!!

To generate HS-CN pairs with the author module, we basically exploited the model test setting and conditioned the fine-tuned model with each HS in the NICHE test-set. 
After removing the CN output for the test HS, we could obtain new pairs of HS-CN. In this way, we generate 2700 HS-CN pairs that we used for our reviewer-configuration experiments.

\section{The Reviewer}

The task of the reviewer % in our scenario is basically 
is a sentence-level Confidence Estimation (CE) %task 
similar to the one of Machine Translation \cite{blatz2004confidence}. In this task, the reviewer must decide whether the author output is correct/suitable for a given source text, i.e. a hate speech. Consistently with the MT scenario, one application of CE is filtering candidates for possible human post-editing, which is conducted by the NGO operator by validating the CN. %that is going to validate the CN. %and, possibly, reordering n-best lists output. 
We tested three reviewer configurations:
%\textbf{(R1)} \textbf{(R2)}\textbf{(R3)}  We didn't use the abbreviations afterwards..
\begin{enumerate}
\itemsep0em 
\item\textbf{expert-reviewer}: Author output is directly presented to NGO operators.
\item\textbf{non-expert-reviewer}: Author output is filtered by human reviewers, then validated by operators.
\item\textbf{machine-reviewer}: Filtering is done by a classifier neural-architecture before operator validation.
\end{enumerate}
\subsection{Human Reviewer Experiment}
\label{Human-Reviewers}
In this section we describe the annotation procedure for the non-expert reviewer configuration.

\noindent\textbf{Setup.} We administered the generated 2700 HS-CN pairs %(generated by the author module) 
to three non-expert annotators, and instructed them to evaluate each pair in terms of CN `suitableness' with regard to the corresponding hate speech. %Annotators were gathered in our premises to perform the task.

\noindent\textbf{Instructions.} We briefly described what is an appropriate and suitable CN, %what an appropriate CN is %(de-escalating, challenging the argument and not the person, etc.).
then we instructed them not to overthink during evaluation, but to give a score based on their intuition. We also provided a list of 20 HS-CN pairs exemplifying the proper evaluation.  %that we discussed together to let possible doubts emerge. 

\noindent\textbf{Measurement.} We opted for a scale of 0-3, rather than a %simple
CE binary response, since it allows us to study various thresholds for better data selection. In particular, the meanings of the scores are as follows: 0 is not suitable; 1 is %almost 
suitable with small modifications, such as grammar or semantic; 2 is suitable; and 3 is extremely good %(fantastic)
as a CN. We also ask to discard the pairs in which the hate speech was not well formed.
For each pair we gathered two annotator scores.

\textbf{Filtered Data.} After the non-expert evaluation, we applied two different thresholds to obtain the pairs to be presented to the expert operators: (i) at least a score of 2 by both annotators (Reviewer$_{\geq2}$) yielding high quality data where no post editing is necessary, (ii) at least a score 1 by both annotators (Reviewer$_{\geq1}$) providing reasonable quality with a possible need for post-editing.
%To filter data we applied two different approaches: in one we gathered all pairs that obtained at least a score 2 by both annotators (Reviewer$_{\geq2}$, high quality where no post editing should be needed), in the second we gathered all pairs that obtained at least a score 1 by both annotators (Reviewer$_{\geq1}$, reasonable quality pairs, including those of the previous group, some post editing can be needed).

%\textcolor{red}{Since annotators gave a score every 35 seconds (1.7 judgements per minute) on average, and we required two judgments per pair, in total it was needed 70 second/person to obtain a final judgment for each pair. %Of course the more stringent the filtering criteria, the higher the time to obtain a filtered pair. So, for example, to obtain a single pair with at least a score 2 by both annotators, 700 seconds (almost 12 minutes) are needed on average.}

The statistics reported in Table \ref{tab:annotation_statistics} show that high quality pairs (Reviewer$_{\geq2}$) account for only a small fraction (10\%) of the produced data and only %one out of three 
one third was of reasonable quality (Reviewer$_{\geq1}$), while the vast majority was discarded. Some randomly selected filtered pairs are provided in Appendix. 

%\begin{table}[h!]
%  \centering
%  \begin{tabular}{l r r r}
%    \hline
%   Threshold & count & Percentage & seconds*pair \\
%    \hline 
%both  $\geq$ 2	& 276	& 10.0\% & 703 \\
%both $\geq$ 1	& 902	& 32.6\% & 215\\
%at least one 0 	& 1723	& 62.2\% & 113 \\
%discarded pairs & 145	& 5.2\% & -\\
%    \hline
%  \end{tabular}
%  \caption{blablabla}\label{tab:annotation_statistics}
%\end{table}

\begin{table}[h!]
  \centering
  \begin{tabular}{l r r }
    \hline
    Threshold & count & Percentage \\
    \hline 
Reviewer$_{\geq2}$	& 276	& 10.0\% \\
Reviewer$_{\geq1}$	& 902	& 32.6\% \\
at least one 0 	& 1723	& 62.2\% \\
bad HS & 145	& 5.2\% \\
    \hline
Reviewer$_{machine}$ & - & 40.2\% \\ %341
    \hline
  \end{tabular}
  \caption{Percentage of filtered pairs according to various filtering conditions.}\label{tab:annotation_statistics}

\end{table}

\subsection{Machine Reviewer Experiment}

%Since the reviewing task is a binary classification problem of assessing the correctness of an NLP systemÕs output (confidence estimation), 
As the machine reviewer we implemented 2 neural classifiers tasked with assessing whether the given HS-CN is a proper data pair. The two models are based on BERT~\cite{devlin2019bert} and ALBERT~\cite{lan2019albert} architectures. %are based one on Bert [CIT] and the other on Albert [CIT]. 

\paragraph{Training data.} We created a balanced dataset with 1373 positive and 1373 negative examples for training purposes. The positive pairs come both from NICHE dataset and from the examples annotated in the human reviewer setting (Reviewer$_{\geq2}$). %NICHE examples were added since the '$\geq$2' condition alone, that represent actual GPT-2 output, did not provide a sufficient amount of training. 
The negative pairs consist of the examples annotated in the human reviewer setting, in the `at least one 0' bin. In addition, 50 random HSs from NICHE-training are utilized with verbatim repetition as HS-HS to discourage the same text for both HS and CN in a pair, and 50 random HSs are paired with other random HSs simulating the condition of inappropriate CNs with hateful text. %\textcolor{blue}{50 random HS from NICHE-training split are repeated both for HS and CN for negative to simulate the exactly same text for both HS and CN, and finally, 50 random HS from NICHE-training are paired with other HS to simulate HS-HS pairs.} \textcolor{purple}{This is not very clear to me, but I can't paraphrase it so far :-/} 
%(\textbf{XX} randomly selected) and from %... \textit{the tfidf experiment 0 from both annotators for negative
%50 random HS from conan-training repeated both for HS and CN for negative to discourage the same text for both HS and CN, 50 random HS from conan-training  are paired with other HS simulating HS-HS kind of dialogue}

\noindent\textbf{Test data.} %For testing purposes 
We collected a balanced test set, with 101 positive and 101 negative pairs. Both positive and negative examples are created replicating the non-expert reviewer annotation described in Section \ref{Human-Reviewers} for new CN generation with NICHE test set by using the author model GPT$_{niche}$. %\textcolor{red}{First, for each HS in the test split, we generated an actual GPT-2 output as CN. DID WE TALK ABOUT THIS ANNOTATION???} % \textit{conan-test for positive
%GPT-2 conan-test evaluation positive and negative scores from 
%both labels > 1 positive 
%both labels == 0 negative }

%We also checked the training and test sets to remove possible accidental overlapping examples, but none was found.
\noindent\textbf{Models.} For the first model, we follow the standard sentence-pair classification fine-tuning schema of the original BERT study. First, the input HS-CN is represented as $[CLS]\; HS\_tokens\;[SEP]\;CN\_tokens\;[SEP]$ and fed into BERT. By using the final hidden state of the first token [CLS] as the input, originally denoted as $C \in R^H$, we %Then, the input sequence is fed into BERT to 
obtain a fixed-dimensional pooled representation of the input sequence. Then, a classification layer is added with the parameter matrix $W \in R^{K?H}$, where K denotes the number of labels, i.e. 2 for HS-CN classification. 
The cross-entropy loss has been used during the fine-tuning.% Per example loss is calculated by taking the negative of the reduce sum of the multiplication of the one hot encoded vector for labels with the $log\_softmax$ output. Then, we calculate the loss of a batch by taking the mean of the per example losses.

We have conducted a hyperparameter tuning phase with a grid-search over the batch sizes 16 and 32, the learning rates [4,3,2,1]e-5 and the number of epochs in the range of 3 to 8. We obtained the best model by fine-tuning uncased BERT-large, with a learning rate of 1e-5, batch size of 16, and after 6 epochs at the $1029^{th}$ step on a single GPU.

The second model is built by fine-tuning ALBERT, which shows better performance than BERT on inter-sentence coherence prediction by using a sentence-order prediction  loss instead of next-sentence prediction. %used for BERT, which conflates topic prediction and coherence prediction.
In sentence-order prediction loss, while the positive examples are created similar to BERT by using the consecutive sentences within the same document, the negative examples are created by swapping sentences, which leads the model to capture the discourse-level coherence properties better \cite{lan2019albert}. % finer-grained distinctions about discourse-level coherence properties . 
This objective is particularly suitable for HS-CN pair classification task, since HS and CN order and their coherence %between them 
are crucial for our task. %. %as well as topic prediction among the HS and CN in which ALBERT also shows a reasonable performance \cite{lan2019albert}. 
We fine-tuned ALBERT similarly to BERT model, by adding a classification layer after the last hidden layer. %of the pretrained model. 
%We performed a grid-search over the batch sizes 16 and 32, learning rates 4e-5, 3e-5, 2e-5 and 1e-5, epochs in the range of 3 to 8,
We applied the same grid-search %procedure
that we used for BERT model 
to fine-tune ALBERT-xxlarge which contains 235M parameters. We saved a checkpoint at every 200 steps and finally, obtained the best model by using the learning rate of 1e-5, the batch size of 16, and at the $1200^{th}$ step.\footnote{All the experiments have been conducted on a single GeForce RTX 2080 Ti GPU. Only the ALBERT classifier model has been trained with 8 TPU cores on Google Cloud.}  

\noindent\textbf{Metrics}. %During the evaluation of the performances of both BERT and ALBERT models, we have used the test set instead of a left-out development set, since i) we have a very limited dataset, ii) we conduct a human evaluation phase as the final performance evaluation. 
To find the best model for machine reviewer, we compared %all configurations of 
BERT and ALBERT models over the test set. Although it seems more intuitive to focus on precision since we search for an effective filtering over many possible solutions, we observed that a model with a very high precision tends to overfit on generic responses, such as ``\textit{Evidence please?}". %, which can be found in the training set more than once and can be applied to many cases. 
Therefore, we aim to keep the balance between the precision and recall and we opted for F1 score for model selection. We report the best configurations for each model in Table \ref{tab:machine_reviewer}, and the percentage of filtered pairs in Table \ref{tab:annotation_statistics}. ALBERT classifier overperformed BERT model in all three metrics; F1, Precision, and Recall. Considering 6\% of absolute F1 score improvement with respect to BERT model, we employed ALBERT model as the Machine Reviewer. %RR (diversity) is also used to make sure that the filtering process do not hamper the variability of the output. 

%Best Configurations for Albert and Bert Results:
%f1 alBert xxLarge 1200 1e05 16 xxlarge 0.7329843 
%Trained on: 8 TPU

%f1 bert large uncased  batch 16 e1-05 epoch 6 1029 %0.6700507
%Trained on: single gpu
%NOTE TO SERRA(MYSELF) run filtering using bert.
\begin{table}[h!]
  \centering
  \begin{tabular}{l r r r}
    \hline
    Reviewer$_{machine}$ & F1 & Precision & Recall\\
    \hline 
ALBERT  & 0.73	& 0.74 & 0.73 \\
BERT & 	0.67 & 0.69 & 0.65 \\
    \hline
  \end{tabular}
  \caption{F1, Precision and Recall results for the two main classifier configurations we tested.% \textbf{Let's see if to keep:}  RR (diversity) is reported to make sure that the filtering process do not hamper the variability of the output.
  }\label{tab:machine_reviewer}

\end{table}

%REWRITE: Albert gave us a boost in the performance. So we used that as the reviewer.

%\section{Training with new obtained data} 
\begin{table*}[h]
  \centering
  \begin{tabular}{l r r | r r | r r}
    \hline
    Approach & NGO$_{time}$ & Crowd$_{time}$ & RR & Novelty & Pairs$_{selec}$ & Pairs$_{final}$\\
    \hline 
    no suggestion & 480 & - & 2.72 & - & - & - \\
    Reviewer$_{expert}$ & 76 & - & 3.56 & 0.73 & 100\% & 45\% \\ %GPT-2$_{no\_filter}$
    Reviewer$_{\geq1}$ & 72 & 215 & 4.31 & 0.70 & 33\% & 54\% \\  %GPT-2$_{filter \geq1}$   
    Reviewer$_{machine}$  & 68 & - & 4.48 & 0.68 & 40\% & 63\% \\
    Reviewer$_{\geq2}$ & 49 & 703 & 5.70 & 0.65 & 10\% & 72\% \\ %GPT-2$_{filter \geq2}$
     %GPT-2$_{automatic\_filter}$
    \hline
  \end{tabular}
  \caption{Results for CN collection under various configurations. RR for `no suggestion' is computed on NICHE dataset and the time needed is the one reported in \cite{conan-2019}. Time is expressed in seconds. Pairs$_{selec}$ indicates the percentage of original author pairs that have been passed to the expert for reviewing, Pairs$_{final}$ indicates the percentage of selected pairs that have been accepted or modified by the expert themselves. Crowd$_{time}$ is computed considering that annotators gave a score every 35 seconds, and we required two judgments per  pair.}\label{tab:time_table1}
\end{table*}

%\section{NGO's operators experiments}
\section{NGO Operators Experiments}
%In this section we present a series of experiments to understand if serving pre-produced counter-narrative suggestions using our author-reviewer approach can actually help data collection. In particular we want to understand if modification/validation of counter narratives is a less time-consuming activity for NGO operators, as compared to manually writing them from scratch. 
To verify that the author-reviewer approach can boost HS-CN data collection, we run an experiment with 5 expert operators from an NGO. 
We compared the filtering strategies to reveal the best depending on several metrics constraints. 
%(filtering conditions)to see which is the best according to given constraints. 
%For this reason, we kept the logs of operator activities during the deployment days, and in particular we recorded the time needed in writing new responses per CN suggestion. The assumption is that with the CN suggestion tool the time needed for composing new or modifying responses would be shorter compared to the one without suggestion tool.

%Therefore we recorded how many suggested messages were further modified and used by operators, and how many were written by operators and saved through the platform after seeing the CN suggestions. 

%In total, using the platform we collected 999 modified messages and 335 new counter-narratives across three countries. 
						
\noindent\textbf{Within Subject Design.} %The evaluation task used a within subject design.
We administered lists of HS-CN pairs to 5 operators from each filtering condition, and instructed them to evaluate/modify each pair in terms of `suitableness' of the CN %with regard 
to the corresponding HS.

\noindent\textbf{Instructions.} For each HS-CN pair we asked the operators: 
a) if the CN is a Òperfect answerÓ, % we asked them
to validate it without any modification. 							
b) if the CN is not perfect, but a good answer can be obtained with some editing, to modify it.
%if the suggestion is not so perfect, but by editing it a good answer can be reasonably obtained, we asked them to modify the CN to make it suitable.							
c) if the CN is completely irrelevant and/or needs to be completely rewritten to fit the given HS, to discard it.%then we asked the operator to discard it. 		

%\begin{table*}[ht!]
% \centering
%  \begin{tabular}{l r r | r r r r | r | r}
%    \hline
%    Approach & NGO$_{time}$ & Crowd$_{time}$ & Div.$_{bef}$  & Div.$_{aft}$ & Nov.$_{bef}$ & Nov.$_{aft}$ & \textbf{TER} & Final-Pairs\\
%    \hline 
%    no suggestion & 480 & - & 2.72 & - & - & - & - & - \\
%    %IR\_suggestion & 300 & - & \textcolor{red}{Yiling} & - & - & - & - & - \\
%    % IR\_suggestion\_new & ?? & - & \\
%    Reviewer$_{expert}$ & 76 & - & 3.56 & 4.33 & \textbf{0.73} & 0.87 & 0.20 & 0.45 \\ %GPT-2$_{no\_filter}$
%    Reviewer$_{\geq1}$ & 72 & 215 & 4.31 & \textcolor{red}{5.73 ??} & \textbf{0.70} & 0.82 & 0.13 & 0.54 \\  %GPT-2$_{filter \geq1}$   
%    Reviewer$_{machine}$  & 68 & - & 4.48 & \textcolor{red}{7.36 ??} & \textbf{0.68} & 0.82 & 0.19 & 0.63 \\
%    Reviewer$_{\geq2}$ & 49 & 703 & 5.70 & 5.51 & \textbf{0.65} & 0.75 & 0.10 & 0.72 \\ %GPT-2$_{filter \geq2}$
%     %GPT-2$_{automatic\_filter}$
%    \hline
%  \end{tabular}
%  \caption{\textcolor{blue}{Time required for writing CN under various configurations. For the condition `no suggestion' the diversity is the one of NICHE  dataset itself. Time is expressed in seconds. \textcolor{red}{"before" is "CN\_suggestion" column, "after" is "your answer" column. USE ONLY YOUR ANSWER }}}\label{tab:time_table1}
%\end{table*}

\noindent\textbf{Measurement.} The main goal of our effort is to reduce the time needed by experts 
to produce training data for automatic CN generation. Therefore the primary evaluation measure is the average time needed to obtain a proper pair. % from each condition. %Then 
The other measurements of interest are Diversity and Novelty, to understand how the reviewing procedure can affect the variability of the obtained pairs.

\noindent\textbf{Procedure and material.} %Again, we wanted to keep the task as simple as possible, so that the operator speed is very high (if easily modifiable do it otherwise go on with the next pair). 
We gave the %aforementioned 
instructions %plus 
along with a list of 20 HS-CN exemplar pairs for each condition (i.e. Reviewer$_{\geq1}$, $_{\geq2}$, $_{machine}$, $_{expert}$). The condition order was randomized to avoid primacy effect. In total, each NGO operator evaluated 80 pairs. Pairs were sampled from the pool of 2700 pairs described before (apart from %those in 
the automatic filtering condition). To guarantee that the sample was representative of the corresponding condition, we performed a stratified sampling and avoided repeating pairs across subjects.
%To grant that the sample was representative of the corresponding condition, we performed a stratified sampling and avoid that pairs are repeated across subjects. 
%\textbf{Subjects.} We hired \textbf{4} NGO operators for the task. These operators come from an NGO that is actively pursuing CN policies to fight HS online. \\

\noindent\textbf{Results and Discussion.} As it is shown in Table \ref{tab:time_table1}, there is a substantial decrease in data collection time (NGO$_{time}$) when automatic generation mechanisms are introduced (no suggestion vs. Reviewer$_{expert}$). If crowd filtering is applied (Reviewer$_{\geq1}$, $_{\geq2}$), the amount of time can be further reduced, and the more stringent the filtering criterion, the higher the time saved. Conversely, the more stringent the filtering criterion, the higher the time to obtain a filtered pair from non-expert annotators (CROWD$_{time}$). %So, for example, 
For instance to obtain a single pair with at least a score of 2 by both annotators, 700 sec (around 12 min) are needed on average (only 10\% of examples are in $\geq2$ condition). %These results all
Results indicate that %the idea of 
providing an automatic generation tool meets the first goal of increasing efficiency of the operators in data collection. 

Regarding diversity and novelty metrics, % we see that
pre-filtering author's output (Reviewer$_{\geq1}$, $_{\geq2}$ and $_{machine}$) %always 
has a negative impact: the more stringent the filtering condition the higher the RR and the lower the novelty of the filtered CNs. %\textcolor{red}{
We performed some manual analysis of the selected CNs and we observed that especially for the Reviewer$_{\geq2}$ case (which was the most problematic in terms of RR and novelty) % we observed %
there was a significantly higher ratio of ``generic" responses, such as ``\textit{This is not true.}" or ``\textit{How can you say this about an entire faith?}" , for which reviewersÕ agreement is easier to attain. Therefore, the higher agreement on the generic CNs reveals itself as a negative impact in the diversity and novelty metrics. %}
Conversely, the percentage of pre-filtered pairs that are accepted by the expert increases with the filtering condition becoming more stringent, the baseline being 45\% for Reviewer$_{expert}$ condition. %(i.e. the pairs accepted if no filtering is provided, Reviewer$_{expert}$ condition).

As for the amount of operators' effort, we observed a slight decrease in HTER\footnote{Human-targeted Translation Edit Rate is a measure of post-editing effort at sentence level translations \cite{specia2010estimating}.} with the increase of pre-filtering conditions, indicating an %increase
improvement in the quality of candidates. %In any case
However, HTER scores were all %was always included 
between 0.1 and 0.2, much below the 0.4 acceptability threshold defined by \citet{turchi2013coping}, indicating that operators %followed the indication of modifying 
modified CNs only if ``easily" amendable. 
Finally, we observe that despite reducing the ouput diversity and novelty, the reduction of expert effort %thanks to the high quality human filtering of silver data
by Reviewer$_{\geq2}$ in terms of the percentage of the obtained pairs is not attainable by a machine yet. On the other hand, automatic filtering (Reviewer$_{machine}$) is a viable solution since (i) it helps the NGO operators save time better than human filter $\geq$1, % (ii) it filters out less pairs than other non-experts configurations??? %Reviewer$_{\geq1}$ (0.33 vs 0.40) while having an higher percentage of pairs accepted (0.54 vs. 0.63),
(ii) it better preserve diversity and novelty as compared to Reviewer$_{\geq2}$ and in line with Reviewer$_{\geq1}$.

%\textcolor{blue}{Finally, we observe that the reduction of expert effort thanks to the high quality human filtering of silver data (Reviewer$_{\geq2}$) is not attainable by a machine yet, still this filtering comes at the price of a drastic reduction in ouput diversity and novelty. At the same time automatic filtering (machine reviewer) is a viable solution since (i) it grants a NGO saving time better than human filter $\geq$1, (ii) it filters out less pairs than other non-experts configurations %Reviewer$_{\geq1}$ (0.33 vs 0.40) while having an higher percentage of pairs accepted (0.54 vs. 0.63), 
%(iii) it better preserve diversity and novelty as compared to Reviewer$_{\geq2}$ and in line with Reviewer$_{\geq1}$.}

%\textcolor{violet}{NOTE: for a model pipeline it might take some seconds to generate and filter a sample, lets not forget about this situation.}

\section{Conclusions}

To counter hatred online and avoid the undesired effects that come with content moderation, intervening in the discussion directly with textual responses is considered as a viable solution. In this scenario, automation strategies, such as natural language generation, are necessary to help NGO operators in their countering effort. However, these automation approaches are not mature yet, since they suffer from the lack of sufficient amount of quality data and tend to produce generic/repetitive responses. %Being aware of 
Considering the aforementioned limitations, we presented a study on how to reduce data collection effort, using a mix of several strategies. To effectively and efficiently obtain varied and novel data, we first propose the generation of silver counter-narratives -- using large scale unsupervised language models -- then a filtering stage by crowd-workers and finally an expert validation/post-editing. We also show promising results obtained by replacing crowd-filtering with an automatic classifier. As a final remark, we believe that the proposed framework can be useful for other NLG tasks such as paraphrase generation or text simplification. 

\section*{Acknowledgments}
This work was partly supported by the HATEMETER project within the EU Rights, Equality and Citizenship Programme 2014-2020.  
We are grateful to Stop Hate UK that provided us with the experts for the evaluation.
Finally, there are also many people we would like to thank for their help and useful suggestions: Eneko Agirre, Simone Magnolini, Marco Turchi, Sara Tonelli and the anonymous reviewers among others.

\bibliographystyle{acl_natbib}

\begin{thebibliography}{51}
\expandafter\ifx\csname natexlab\endcsname\relax\def\natexlab#1{#1}\fi

\bibitem[{Benesch(2014)}]{benesch2014countering}
Susan Benesch. 2014.
\newblock Countering dangerous speech: New ideas for genocide prevention.
\newblock \emph{Washington, DC: United States Holocaust Memorial Museum}.

\bibitem[{Benesch et~al.(2016)Benesch, Ruths, P~Dillon, Mohammad~Saleem, and
  Wright}]{benesch2016}
Susan Benesch, Derek Ruths, Kelly P~Dillon, Haji Mohammad~Saleem, and Lucas
  Wright. 2016.
\newblock Counterspeech on twitter: A field study.
\newblock \emph{Dangerous Speech Project. Available at:
  https://dangerousspeech.org/counterspeech-on-twitter-a-field- study/}.

\bibitem[{Bertoldi et~al.(2013)Bertoldi, Cettolo, and
  Federico}]{bertoldi2013cache}
Nicola Bertoldi, Mauro Cettolo, and Marcello Federico. 2013.
\newblock Cache-based online adaptation for machine translation enhanced
  computer assisted translation.
\newblock In \emph{MT-Summit}, pages 35--42.

\bibitem[{Bielefeldt et~al.(2011)Bielefeldt, La~Rue, and
  Muigai}]{Bielefeldt2011}
Heiner Bielefeldt, Frank La~Rue, and Githu Muigai. 2011.
\newblock Ohchr expert workshops on the prohibition of incitement to national,
  racial or religious hatred.
\newblock In \emph{Expert workshop on the Americas}.

\bibitem[{Blatz et~al.(2004)Blatz, Fitzgerald, Foster, Gandrabur, Goutte,
  Kulesza, Sanchis, and Ueffing}]{blatz2004confidence}
John Blatz, Erin Fitzgerald, George Foster, Simona Gandrabur, Cyril Goutte,
  Alex Kulesza, Alberto Sanchis, and Nicola Ueffing. 2004.
\newblock Confidence estimation for machine translation.
\newblock In \emph{Coling 2004: Proceedings of the 20th international
  conference on computational linguistics}, pages 315--321.

\bibitem[{Burnap and Williams(2015)}]{burnap2015cyber}
Pete Burnap and Matthew~L Williams. 2015.
\newblock Cyber hate speech on twitter: An application of machine
  classification and statistical modeling for policy and decision making.
\newblock \emph{Policy \& Internet}, 7(2):223--242.

\bibitem[{Burnap and Williams(2016)}]{burnap2016us}
Pete Burnap and Matthew~L Williams. 2016.
\newblock Us and them: identifying cyber hate on twitter across multiple
  protected characteristics.
\newblock \emph{EPJ Data Science}, 5(1):11.

\bibitem[{Cettolo et~al.(2014)Cettolo, Bertoldi, and
  Federico}]{cettolo2014repetition}
Mauro Cettolo, Nicola Bertoldi, and Marcello Federico. 2014.
\newblock The repetition rate of text as a predictor of the effectiveness of
  machine translation adaptation.
\newblock In \emph{Proceedings of the 11th Biennial Conference of the
  Association for Machine Translation in the Americas (AMTA 2014)}, pages
  166--179.

\bibitem[{Chung et~al.(2019)Chung, Kuzmenko, Tekiroglu, and
  Guerini}]{conan-2019}
Yi-Ling Chung, Elizaveta Kuzmenko, Serra~Sinem Tekiroglu, and Marco Guerini.
  2019.
\newblock \href {https://doi.org/10.18653/v1/P19-1271} {{CONAN} - {CO}unter
  {NA}rratives through nichesourcing: a multilingual dataset of responses to
  fight online hate speech}.
\newblock In \emph{Proceedings of the 57th Annual Meeting of the Association
  for Computational Linguistics}, pages 2819--2829, Florence, Italy.
  Association for Computational Linguistics.

\bibitem[{Del~Vigna et~al.(2017)Del~Vigna, Cimino, Dell’Orletta, Petrocchi,
  and Tesconi}]{del2017hate}
Fabio Del~Vigna, Andrea Cimino, Felice Dell’Orletta, Marinella Petrocchi, and
  Maurizio Tesconi. 2017.
\newblock Hate me, hate me not: Hate speech detection on facebook.

\bibitem[{Devlin et~al.(2019)Devlin, Chang, Lee, and
  Toutanova}]{devlin2019bert}
Jacob Devlin, Ming-Wei Chang, Kenton Lee, and Kristina Toutanova. 2019.
\newblock Bert: Pre-training of deep bidirectional transformers for language
  understanding.
\newblock In \emph{Proceedings of the 2019 Conference of the North American
  Chapter of the Association for Computational Linguistics: Human Language
  Technologies, Volume 1 (Long and Short Papers)}, pages 4171--4186.

\bibitem[{Dziri et~al.(2018)Dziri, Kamalloo, Mathewson, and
  Zaiane}]{dziri2018augmenting}
Nouha Dziri, Ehsan Kamalloo, Kory~W Mathewson, and Osmar Zaiane. 2018.
\newblock Augmenting neural response generation with context-aware topical
  attention.
\newblock \emph{arXiv preprint arXiv:1811.01063}.

\bibitem[{Fortuna and Nunes(2018)}]{fortuna2018survey}
Paula Fortuna and S{\'e}rgio Nunes. 2018.
\newblock A survey on automatic detection of hate speech in text.
\newblock \emph{ACM Computing Surveys (CSUR)}, 51(4):85.

\bibitem[{de~Gibert et~al.(2018)de~Gibert, Perez, Garc{\i}a-Pablos, and
  Cuadros}]{de2018hate}
Ona de~Gibert, Naiara Perez, Aitor Garc{\i}a-Pablos, and Montse Cuadros. 2018.
\newblock Hate speech dataset from a white supremacy forum.
\newblock \emph{EMNLP 2018}, page~11.

\bibitem[{Gitari et~al.(2015)Gitari, Zuping, Damien, and
  Long}]{gitari2015lexicon}
Njagi~Dennis Gitari, Zhang Zuping, Hanyurwimfura Damien, and Jun Long. 2015.
\newblock A lexicon-based approach for hate speech detection.
\newblock \emph{International Journal of Multimedia and Ubiquitous
  Engineering}, 10(4):215--230.

\bibitem[{Golovanov et~al.(2019)Golovanov, Kurbanov, Nikolenko, Truskovskyi,
  Tselousov, and Wolf}]{golovanov2019large}
Sergey Golovanov, Rauf Kurbanov, Sergey Nikolenko, Kyryl Truskovskyi, Alexander
  Tselousov, and Thomas Wolf. 2019.
\newblock Large-scale transfer learning for natural language generation.
\newblock In \emph{Proceedings of the 57th Annual Meeting of the Association
  for Computational Linguistics}, pages 6053--6058.

\bibitem[{Holtzman et~al.(2019)Holtzman, Buys, Forbes, and
  Choi}]{holtzman2019curious}
Ari Holtzman, Jan Buys, Maxwell Forbes, and Yejin Choi. 2019.
\newblock The curious case of neural text degeneration.
\newblock \emph{arXiv preprint arXiv:1904.09751}.

\bibitem[{Hosseinmardi et~al.(2015)Hosseinmardi, Mattson, Rafiq, Han, Lv, and
  Mishra}]{hosseinmardi2015detection}
Homa Hosseinmardi, Sabrina~Arredondo Mattson, Rahat~Ibn Rafiq, Richard Han, Qin
  Lv, and Shivakant Mishra. 2015.
\newblock Detection of cyberbullying incidents on the instagram social network.
\newblock \emph{arXiv preprint arXiv:1503.03909}.

\bibitem[{Hua et~al.(2019)Hua, Hu, and Wang}]{hua2019argument}
Xinyu Hua, Zhe Hu, and Lu~Wang. 2019.
\newblock Argument generation with retrieval, planning, and realization.
\newblock \emph{arXiv preprint arXiv:1906.03717}.

\bibitem[{Jurgens et~al.(2019)Jurgens, Hemphill, and
  Chandrasekharan}]{jurgens-etal-2019-just}
David Jurgens, Libby Hemphill, and Eshwar Chandrasekharan. 2019.
\newblock \href {https://doi.org/10.18653/v1/P19-1357} {A just and
  comprehensive strategy for using {NLP} to address online abuse}.
\newblock In \emph{Proceedings of the 57th Annual Meeting of the Association
  for Computational Linguistics}, pages 3658--3666, Florence, Italy.
  Association for Computational Linguistics.

\bibitem[{Klubi{\v{c}}ka and Fern{\'a}ndez(2018)}]{Klubicka2018}
Filip Klubi{\v{c}}ka and Raquel Fern{\'a}ndez. 2018.
\newblock Examining a hate speech corpus for hate speech detection and
  popularity prediction.
\newblock In \emph{LREC}.

\bibitem[{Kumar et~al.(2018)Kumar, Ojha, Malmasi, and
  Zampieri}]{kumar2018benchmarking}
Ritesh Kumar, Atul~Kr Ojha, Shervin Malmasi, and Marcos Zampieri. 2018.
\newblock Benchmarking aggression identification in social media.
\newblock In \emph{Proceedings of the First Workshop on Trolling, Aggression
  and Cyberbullying (TRAC-2018)}, pages 1--11.

\bibitem[{Lan et~al.(2019)Lan, Chen, Goodman, Gimpel, Sharma, and
  Soricut}]{lan2019albert}
Zhenzhong Lan, Mingda Chen, Sebastian Goodman, Kevin Gimpel, Piyush Sharma, and
  Radu Soricut. 2019.
\newblock Albert: A lite bert for self-supervised learning of language
  representations.
\newblock \emph{arXiv preprint arXiv:1909.11942}.

\bibitem[{Li et~al.(2016)Li, Galley, Brockett, Gao, and
  Dolan}]{li-etal-2016-diversity}
Jiwei Li, Michel Galley, Chris Brockett, Jianfeng Gao, and Bill Dolan. 2016.
\newblock \href {https://doi.org/10.18653/v1/N16-1014} {A diversity-promoting
  objective function for neural conversation models}.
\newblock In \emph{Proceedings of the 2016 Conference of the North {A}merican
  Chapter of the Association for Computational Linguistics: Human Language
  Technologies}, pages 110--119, San Diego, California. Association for
  Computational Linguistics.

\bibitem[{Manurung et~al.(2008)Manurung, Ritchie, and
  Thompson}]{manurung-etal-2008-implementation}
Ruli Manurung, Graeme Ritchie, and Henry Thompson. 2008.
\newblock \href {https://www.aclweb.org/anthology/Y08-1027} {An implementation
  of a flexible author-reviewer model of generation using genetic algorithms}.
\newblock In \emph{Proceedings of the 22nd Pacific Asia Conference on Language,
  Information and Computation}, pages 272--281, The University of the
  Philippines Visayas Cebu College, Cebu City, Philippines. De La Salle
  University, Manila, Philippines.

\bibitem[{Mathew et~al.(2018)Mathew, Kumar, Goyal, Mukherjee
  et~al.}]{mathew2018analyzing}
Binny Mathew, Navish Kumar, Pawan Goyal, Animesh Mukherjee, et~al. 2018.
\newblock Analyzing the hate and counter speech accounts on twitter.
\newblock \emph{arXiv preprint arXiv:1812.02712}.

\bibitem[{Mathew et~al.(2019)Mathew, Saha, Tharad, Rajgaria, Singhania, Maity,
  Goyal, and Mukherjee}]{mathew2019thou}
Binny Mathew, Punyajoy Saha, Hardik Tharad, Subham Rajgaria, Prajwal Singhania,
  Suman~Kalyan Maity, Pawan Goyal, and Animesh Mukherjee. 2019.
\newblock Thou shalt not hate: Countering online hate speech.
\newblock In \emph{Proceedings of the International AAAI Conference on Web and
  Social Media}, volume~13, pages 369--380.

\bibitem[{Munger(2017)}]{munger2017tweetment}
Kevin Munger. 2017.
\newblock Tweetment effects on the tweeted: Experimentally reducing racist
  harassment.
\newblock \emph{Political Behavior}, 39(3):629--649.

\bibitem[{Oberlander and Brew(2000)}]{oberlander2000stochastic}
Jon Oberlander and Chris Brew. 2000.
\newblock Stochastic text generation.
\newblock \emph{Philosophical Transactions of the Royal Society of London.
  Series A: Mathematical, Physical and Engineering Sciences},
  358(1769):1373--1387.

\bibitem[{Papineni et~al.(2002)Papineni, Roukos, Ward, and
  Zhu}]{papineni2002bleu}
Kishore Papineni, Salim Roukos, Todd Ward, and Wei-Jing Zhu. 2002.
\newblock Bleu: a method for automatic evaluation of machine translation.
\newblock In \emph{Proceedings of the 40th annual meeting on association for
  computational linguistics}, pages 311--318. Association for Computational
  Linguistics.

\bibitem[{Qian et~al.(2019)Qian, Bethke, Liu, Belding, and
  Wang}]{qian-etal-2019-benchmark}
Jing Qian, Anna Bethke, Yinyin Liu, Elizabeth Belding, and William~Yang Wang.
  2019.
\newblock A benchmark dataset for learning to intervene in online hate speech.
\newblock In \emph{Proceedings of the 2019 Conference on Empirical Methods in
  Natural Language Processing and the 9th International Joint Conference on
  Natural Language Processing (EMNLP-IJCNLP)}, pages 4757--4766, Hong Kong,
  China. Association for Computational Linguistics.

\bibitem[{Radford et~al.(2019)Radford, Wu, Child, Luan, Amodei, and
  Sutskever}]{radford2019language}
Alec Radford, Jeffrey Wu, Rewon Child, David Luan, Dario Amodei, and Ilya
  Sutskever. 2019.
\newblock Language models are unsupervised multitask learners.
\newblock \emph{OpenAI Blog}, 1(8).

\bibitem[{Richards(1987)}]{richards1987type}
Brian Richards. 1987.
\newblock Type/token ratios: What do they really tell us?
\newblock \emph{Journal of child language}, 14(2):201--209.

\bibitem[{Ross et~al.(2017)Ross, Rist, Carbonell, Cabrera, Kurowsky, and
  Wojatzki}]{ross2017measuring}
Bj{\"o}rn Ross, Michael Rist, Guillermo Carbonell, Benjamin Cabrera, Nils
  Kurowsky, and Michael Wojatzki. 2017.
\newblock Measuring the reliability of hate speech annotations: The case of the
  european refugee crisis.
\newblock \emph{arXiv preprint arXiv:1701.08118}.

\bibitem[{Schieb and Preuss(2016)}]{schieb2016governing}
Carla Schieb and Mike Preuss. 2016.
\newblock Governing hate speech by means of counterspeech on facebook.
\newblock In \emph{66th ica annual conference, at fukuoka, japan}, pages 1--23.

\bibitem[{Schmidt and Wiegand(2017)}]{schmidt2017survey}
Anna Schmidt and Michael Wiegand. 2017.
\newblock A survey on hate speech detection using natural language processing.
\newblock In \emph{Proceedings of the Fifth International Workshop on Natural
  Language Processing for Social Media}, pages 1--10.

\bibitem[{Silva et~al.(2016)Silva, Mondal, Correa, Benevenuto, and
  Weber}]{silva2016analyzing}
Leandro~Ara{\'u}jo Silva, Mainack Mondal, Denzil Correa, Fabr{\'\i}cio
  Benevenuto, and Ingmar Weber. 2016.
\newblock Analyzing the targets of hate in online social media.
\newblock In \emph{ICWSM}, pages 687--690.

\bibitem[{Silverman et~al.(2016)Silverman, Stewart, Birdwell, and
  Amanullah}]{silverman2016impact}
Tanya Silverman, Christopher~J Stewart, Jonathan Birdwell, and Zahed Amanullah.
  2016.
\newblock The impact of counter-narratives.
\newblock \emph{Institute for Strategic Dialogue, London. https://www.
  strategicdialogue.
  org/wp-content/uploads/2016/08/Impact-of-Counter-Narratives\_ONLINE.
  pdf--73}.

\bibitem[{Specia and Farzindar(2010)}]{specia2010estimating}
Lucia Specia and Atefeh Farzindar. 2010.
\newblock Estimating machine translation post-editing effort with hter.
\newblock In \emph{Proceedings of the Second Joint EM+/CNGL Workshop Bringing
  MT to the User: Research on Integrating MT in the Translation Industry (JEC
  10)}, pages 33--41.

\bibitem[{Sprugnoli et~al.(2018)Sprugnoli, Menini, Tonelli, Oncini, and
  Piras}]{sprugnoli2018creating}
Rachele Sprugnoli, Stefano Menini, Sara Tonelli, Filippo Oncini, and Enrico
  Piras. 2018.
\newblock Creating a whatsapp dataset to study pre-teen cyberbullying.
\newblock In \emph{Proceedings of the 2nd Workshop on Abusive Language Online
  (ALW2)}, pages 51--59.

\bibitem[{Stroud and Cox(2018)}]{stroud2018varieties}
Scott~R Stroud and William Cox. 2018.
\newblock The varieties of feminist counterspeech in the misogynistic online
  world.
\newblock In \emph{Mediating Misogyny}, pages 293--310. Springer.

\bibitem[{Turchi et~al.(2013)Turchi, Negri, and Federico}]{turchi2013coping}
Marco Turchi, Matteo Negri, and Marcello Federico. 2013.
\newblock Coping with the subjectivity of human judgements in mt quality
  estimation.
\newblock In \emph{Proceedings of the Eighth Workshop on Statistical Machine
  Translation}, pages 240--251.

\bibitem[{Vaswani et~al.(2017)Vaswani, Shazeer, Parmar, Uszkoreit, Jones,
  Gomez, Kaiser, and Polosukhin}]{vaswani2017attention}
Ashish Vaswani, Noam Shazeer, Niki Parmar, Jakob Uszkoreit, Llion Jones,
  Aidan~N Gomez, {\L}ukasz Kaiser, and Illia Polosukhin. 2017.
\newblock Attention is all you need.
\newblock In \emph{Advances in neural information processing systems}, pages
  5998--6008.

\bibitem[{Wang and Wan(2018)}]{wang2018sentigan}
Ke~Wang and Xiaojun Wan. 2018.
\newblock Sentigan: Generating sentimental texts via mixture adversarial
  networks.
\newblock In \emph{IJCAI}, pages 4446--4452.

\bibitem[{Warner and Hirschberg(2012)}]{warner2012detecting}
William Warner and Julia Hirschberg. 2012.
\newblock Detecting hate speech on the world wide web.
\newblock In \emph{Proceedings of the Second Workshop on Language in Social
  Media}, pages 19--26. Association for Computational Linguistics.

\bibitem[{Waseem(2016)}]{waseem2016you}
Zeerak Waseem. 2016.
\newblock Are you a racist or am i seeing things? annotator influence on hate
  speech detection on twitter.
\newblock In \emph{Proceedings of the first workshop on NLP and computational
  social science}, pages 138--142.

\bibitem[{Waseem and Hovy(2016)}]{waseem2016hateful}
Zeerak Waseem and Dirk Hovy. 2016.
\newblock Hateful symbols or hateful people? predictive features for hate
  speech detection on twitter.
\newblock In \emph{Proceedings of the NAACL student research workshop}, pages
  88--93.

\bibitem[{Wolf et~al.(2019)Wolf, Sanh, Chaumond, and
  Delangue}]{wolf2019transfertransfo}
Thomas Wolf, Victor Sanh, Julien Chaumond, and Clement Delangue. 2019.
\newblock Transfertransfo: A transfer learning approach for neural network
  based conversational agents.
\newblock \emph{arXiv preprint arXiv:1901.08149}.

\bibitem[{Xiang et~al.(2012)Xiang, Fan, Wang, Hong, and
  Rose}]{xiang2012detecting}
Guang Xiang, Bin Fan, Ling Wang, Jason Hong, and Carolyn Rose. 2012.
\newblock Detecting offensive tweets via topical feature discovery over a large
  scale twitter corpus.
\newblock In \emph{Proceedings of the 21st ACM international conference on
  Information and knowledge management}, pages 1980--1984. ACM.

\bibitem[{Xu et~al.(2018)Xu, Ren, Lin, and Sun}]{xu-etal-2018-diversity}
Jingjing Xu, Xuancheng Ren, Junyang Lin, and Xu~Sun. 2018.
\newblock \href {https://doi.org/10.18653/v1/D18-1428} {Diversity-promoting
  {GAN}: A cross-entropy based generative adversarial network for diversified
  text generation}.
\newblock In \emph{Proceedings of the 2018 Conference on Empirical Methods in
  Natural Language Processing}, pages 3940--3949, Brussels, Belgium.
  Association for Computational Linguistics.

\bibitem[{Zhang et~al.(2019)Zhang, Kishore, Wu, Weinberger, and
  Artzi}]{zhang2019bertscore}
Tianyi Zhang, Varsha Kishore, Felix Wu, Kilian~Q Weinberger, and Yoav Artzi.
  2019.
\newblock Bertscore: Evaluating text generation with bert.
\newblock \emph{arXiv preprint arXiv:1904.09675}.

\end{thebibliography}

\FloatBarrier
\clearpage
\appendix

\section{Appendices}
\label{sec:appendix}

\begin{table}[h]
%\begin{centering}
\centering
\begin{tabular}{l|p{13.5cm}}%{lrrr}
     \hline
HS & I love the Pres also he is a real man É not a commie cunt like the last muslim in there. \\
 \hline
 \multirow{3}{*}{GPT2} & Use of the c\-word here is totally unacceptable because it is hurtful and offensive.\\ %it demeans women.\\ 
 \cline{2-2}
  & That is very interesting, but they already have that kind of man who is always saying something different. No need for your comment. \\     
  \cline{2-2} %% here
  & Hi there, please refrain from using derogatory comments in the thread. They are hurtful and unwanted. If you continue Admin will be alerted.  \\
     \hline
\multirow{3}{*}{TRF} & Please refrain from using derogatory terms for females. \\ \cline{2-2}
 & It's really hard to respect someone who uses racial slurs in an attempt at comedy. \\ \cline{2-2}          
 & Please don't use hateful words, please use your words or else removal will take place.  \\    
  
\hline
\hline
HS & What cha doin America? Oh...just waiting to hang the cunt the Muslim nigga. \\
 \hline
\multirow{3}{*}{GPT2} & Name calling speaks volumes about one's character. \\ \cline{2-2}
 & To act like those of a race are the same is to admit to your narrow mindedness. \\ \cline{2-2}          %% here
 & Hey there, racist and derogatory views are hurtful and uncalled for. Please refrain from this type of behavior or you will be reported. Thx.  \\ 
     \hline
\multirow{3}{*}{TRF} & It doesn't make sense to name call someone you've never met based on what the media says. \\ \cline{2-2}
& I'd be more inclined to listen to your side if you didn't use that slur. \\ \cline{2-2}          %% here
& Please muslims are not so two as a high everyone has no need to get an ignorant little elected official further. \\ 
 \hline
\hline
HS & ``Turkish Nazi Muslims" huh retard. Hey, try forming real sentences instead of half of your worlds being fucking hashtags, fuckwit. \\
 \hline
\multirow{3}{*}{GPT2} & It is acceptable for you to talk like this but not okay to physically attack others with it. \\ \cline{2-2}
& Some people find the Nazi Holocaust story appalling, but we should not use those slurs. \\ \cline{2-2}          %% here
& Please refrain from using derogatory terms for the mentally handicapped. \\ 
     \hline
\multirow{3}{*}{TRF} & Please do not use offensive language.\\ \cline{2-2}
  & Use of the r word is unacceptable because it hurts, insults and demeans people with mental disabilities. \\ \cline{2-2}          %% here
 & That word is offensive to the intellectually disabled. Furthermore, there is no shame in being intellectually disabled. \\ 
 \hline
\hline
HS & The muslim nigger terrorist obama needs death the hussein needs it's limbs ripped off and his tongue seared with a hot poker.\\
 \hline
  \multirow{3}{*}{GPT2} & You don't know that, but I do. \\ \cline{2-2}
  & I understand your point but please watch the language. \\ \cline{2-2}          %% here
  & You are throwing a lot of hatred and anti-Semite in to this, which is not good. \\ 
     \hline
\multirow{3}{*}{TRF} & I'd be inclined to agree with you if you didn't use slurs. \\ \cline{2-2}
& Your reply isn't need, not big or clearer. Think before you type next time please. \\ \cline{2-2}          %% here
& All religions, ethnicities, and social groups deserve tolerance. \\ 
     \hline
\end{tabular}
%\end{center}
\caption{\parbox{15cm}{Randomly sampled CNs generated from GPT-2 and TRF models trained on CROWD dataset.}}
  \label{table:ww_examples}
  %\vspace{-4mm}
\end{table}

%\begin{table}[h]
%\begin{center}
%\centering
%\begin{tabular}{l|p{13.5cm}}%{lrrr}
%     \hline
%HS & Islam corrupts our children in schools and this is a shame.\\
% \hline
%\multicolumn{3}{l|}{\multirow{GPT2}} & The truth is that many imams are really committed to education and offer their children in mosques and Christian schools. This is a noble trade. \\ \cline{2-2}
%  \multicolumn{3}{l|}{} & Children in school should be free to think for themselves and do things that adults do not allow themselves, this is a healthy respect for adults. \\ \cline{2-2}          %% here
%  \multicolumn{3}{l|}{} & What are the chances of meeting an intolerant Muslim child in school? I do not think it happens. \\         %% here
%     \hline
%\multicolumn{3}{l|}{\multirow{TRF}} & At most, children are just learning more about different religions. you are using very strong and unnecessary words here. \\ \cline{2-2}
%  \multicolumn{3}{l|}{} & Maybe if you d learned about islam in school you d understand it is not a threat to you and you d be more tolerant of it. \\ \cline{2-2}          %% here
%  \multicolumn{3}{l|}{} & Learning different religions so we can live in the inclusive society is nonsense? i strongly disagree. \\      
%     \hline
%\end{tabular}
%\end{center}
%\caption{\parbox{15cm}{Randomly sampled CNs generated from GPT-2 and TRF models trained on NICHE dataset.}}
%  \label{table:conan_examples1}
%\end{table}

\begin{table*}[ht!]
\centering
\begin{tabular}{l|p{13.5cm}}%{lrrr}
     \hline
HS & White women converted to Islam are the saddest and most selfish persons that you can encounter. They praise happily Islam beliefs and become part of a sisterhood of tollerance and understanding, but they do not know the situation of women in Pakistan and other islamic states. \\
 \hline
 \multirow{3}{*}{GPT2} & Do you know that in many Muslim-majority countries, women's rights are recognized and actively promoted? Laws are passed that guarantee the right to have children, and foreign charities work to ensure that women have access to quality education and health services. \\ \cline{2-2}
  & Are you a believing woman who has known and respected great men like Mohammed, Albert and Winston Churchill. You have forgotten about those who sacrificed so much for your benefit. \\ \cline{2-2}          %% here
  & If white women who love Islam were doing such things why would their profile be so public?\\  
     \hline
 \multirow{3}{*}{TRF} & Honestly, i see around many more muslim women without burqa or veil than women dressed like europeans. \\ \cline{2-2}
 & Are you sure? muslim nobel prize winners would suggest otherwise. \\ \cline{2-2}          %% here
 & Sounds like you are generalizing what a very small percentage of extremists do to an entire religion.\\ 
 
\hline
\hline
HS & We are every day in constant danger, because Muslims can blow us all up. For this fact we should put all Muslims in a hole in the ground.\\
 \hline
 \multirow{3}{*}{GPT2} & If you had said this about all the non-Muslim groups who may espouse hateful or violent views, how would you feel about being labelled a bigot? \\ \cline{2-2}
 & Apart from being offensive and ignorant, is also the belief that Muslims only create problems. In fact, many Muslims are working to solve some of these social problems. \\ \cline{2-2}          %% here
 & Muslims have never sought to destroy our world, rather they have offered their goods and services in return for peace and unity. Nostalgia may have something to offer us, but not for taking over countries and nations. \\
     \hline
 \multirow{3}{*}{TRF} & People who commit offenses are punished by law. Muslims are not all terrorists. muslims are not trying to generalize on their own community, indeed anything. If someone would like you, but you are peaceful muslim countries like which, please. \\ \cline{2-2}
 & No, a person can be immigrants for multiple allegiances and responsibilies. \\ \cline{2-2}          %% here
 & Why do you think that muslim criminals will receive different treatment or even privileges? \\
 
 \hline
\hline
HS & Koran is a book that should be burnt and mosques should be demolished. If we do all these things, all problems in our community will be solved.\\
 \hline
 \multirow{3}{*}{GPT2} & All book burning is prohibited under the Universal Declaration of Human Rights, which says: ' Everyone has the right to book his escape from cruelty and oppression wherever it may be found.'? \\ \cline{2-2}
 & Yes, because this book encourages people to respect for life and to respect God. Glad you like it! \\ \cline{2-2}          %% here
 & Praise be to God that we have these humble requests. Can you give us some figures on the number of people who would like to migrate to your country if you could only send em home?  \\
     \hline
     
 \multirow{3}{*}{TRF} & If you dislike diversity so much , why do not you go off and live in a country where there are no muslims, only people like you? \\ \cline{2-2}
 & Mgm fgm have been in britain for hundreds of years. it is never acceptable but it does not come from just one country or ideology. \\ \cline{2-2}          %% here
 & The idea of blow someone up is itself crazy.\\
 \hline
\end{tabular}
\caption{Randomly sampled CNs generated from GPT-2 and TRF models trained on NICHE dataset.}
  \label{table:conan_examples}
\end{table*}

%%%%%%%%%%%%%%%%%%%%%%%%%

\begin{table*}[h!]
\centering
\begin{tabular}{l|p{11.5cm}}%{lrrr}
     \hline     
HS & Girls and boys are brainwashed by the same people.\\
 \hline
No filter & That's why most religions end well.\\\cline{2-2}
Reviewer$_{\geq1}$ & Mindless Hate and fear mongering divide and conquer.\\\cline{2-2}
Reviewer$_{\geq2}$& Mindless Hate and fear mongering do not help anyone.\\
     \hline
     \hline
HS & Everything will be better if Muslims did not exist. Their only contribute to our society is violence.\\
 \hline
%No filter & Did you know that the greatest child abusers in British history were white men? That is if they weren't SO EXTREMELY LIE!?!?!\\
No filter & Did you know that the greatest child abusers in British history were white men? That is if they weren't SO EXTREMELY LIE!?!?! \\ \cline{2-2}
%\hline
\multirow{3}{*}{Reviewer$_{\geq1}$} & Why would you say that? I would say that all the people using social media and whatever to vent their anger and frustration are clearly not serious about their religion. Some may be, and are angry, but that does not mean they are responsible for anything. \\ \cline{2-2}
 & The world would be a better place without forced baptisms and child rapists. \\ \cline{2-2}      %% here
  & What exactly are you suggesting? \\ \cline{2-2}

Reviewer$_{\geq2}$ & The world would actually be a very dark place without Muslims who contribute a lot to our society. How about a place of peace, tolerance and understanding for all? \\ \cline{2-2}

%Reviewer$_{\geq2}$& The world would actually be a very dark place without Muslims who contribute a lot to our society. How about a place of peace, tolerance and understanding for all?\\
     \hline
     \hline
HS & Europe is civilised, Muslims should not stay there.\\
 \hline
No filter & That is a misunderstanding. On the one hand, Europe is a civilized place, on the other, Muslims cannot excel at anything.\\\cline{2-2}
Reviewer$_{\geq2}$& How can you say that about an entire faith of 1.6 billion people?\\    \cline{2-2}
Reviewer$_{machine}$ & Muslims have been a part of the European society for centuries, they have been a part of its institutions since the middle ages, they have been active in politics during the 19$^{th}$ century and have been crucial in the development of Europe during the last decades.\\
          \hline
     \hline
     HS & Multiculturalism has brought us nothing but disaster.\\
 \hline
Reviewer$_{machine}$ & The multiethnic society has produced many smart and talented people, who have gone on to work in prominent public offices. No one is asking us to sacrifice our culture or our identity for the greater good of the community.\\
\hline
\end{tabular}
\caption{Randomly sampled CNs, generated from GPT-2, filtered from various reviewer configurations.}
  \label{table:ww_examples}
\end{table*}

\end{document}